# Weather Forecasting Error in Solar Energy Forecasting

Hossein Sangrody [1*], Morteza Sarailoo [1], Ning Zhou [1], Nhu Tran [2], Mahdi Motalleb[3], Elham Foruzan [4]

[1] Electrical and Computer Engineering Department, Binghamton University, State University of New York, Binghamton, USA
[2] School of Management, Binghamton University, State University of New York, Binghamton, USA
[3]Department of Electrical Engineering, University of Hawaii, Manoa, Hi, USA
[4]Electrical and Computer Engineering Department, University of Nebraska Lincoln, Lincoln, NE, USA
[*]habdoll1@binghamton.edu

**Abstract:** As renewable distributed energy resources (DERs) penetrate the power grid at an accelerating speed, it is essential for operators to have accurate solar photovoltaic (PV) energy forecasting for efficient operations and planning. Generally, observed weather data are applied in the solar PV generation forecasting model while in practice the energy forecasting is based on forecasted weather data. In this paper, a study on the uncertainty in weather forecasting for the most commonly used weather variables is presented. The forecasted weather data for six days ahead is compared with the observed data and the results of analysis are quantified by statistical metrics. In addition, the most influential weather predictors in energy forecasting model are selected. The performance of historical and observed weather data errors is assessed using a solar PV generation forecasting model. Finally, a sensitivity test is performed to identify the influential weather variables whose accurate values can significantly improve the results of energy forecasting.

## 1. Introduction

Both energy and load forecasting play a critical role in planning, control, and operation of power systems. As renewable energy resources are penetrating the power grid at an accelerating speed, their indispatchability, variability, and uncertainty have presented unprecedented challenges to power grid operations and planning. As a result, accurate forecasting is more vital than before [1] [2] [3].

Most of energy forecasting models are trained using observed weather variables and there are lots of studies which focused on improving their forecasting models with efficient methodologies. However, the trained models are applied using forecasted weather variables [4]. The issue of this practice is that if uncorrelated weather variables or forecasted weather variables with huge errors are entered in the forecasting model, the resulting forecast may not be satisfactory. These errors are more severe when forecast lead time gets longer. Consequently, the efficiency of a forecasting model is unacceptable when the inputs suffer from large errors or include uncorrelated data.

Many studies have been carried out to forecast renewable distributed energy resources' (DERs) generation and many methods have been suggested to improve forecasting models [5] [6] [7]. In [8] an intelligent method is proposed to forecast wind speed and solar radiation based on predictive coding and image processing. In [9], authors provided a survey on using ensemble methods for wind speed/power forecasting and solar irradiance forecasting. They concluded that generally the ensemble forecasting methods surpass other non-ensemble methods. A comprehensive survey study on the latest state-of-the-art in solar energy forecasting was conducted in [10]. This paper discusses motivations, effects of forecast horizon, benefits of regional forecast, origin of inputs, and advantages of probabilistic forecast over deterministic forecast. The output errors of hybrid photovoltaic (PV) power forecasting models were studied in [11]. The authors discussed Least Square Support Vector Machines, Artificial Neural Network, and hybrid statistical model based on Least Square Support Vector Machines with Wavelet Decomposition. They used conventional metrics, such as the root mean square error, mean bias error and mean absolute error (MAE), to evaluate the performance of the different methods. In [12], authors proposed a new method for online forecasting of the output power of PV systems. The proposed approach consists of two stages. In first stage normalized statistical solar powers of a clear sky model are acquired. In second stage an adaptive linear time series approach is utilized to forecast the power output. A novel hybrid algorithm was proposed in [13] to forecast output power of a PV. The proposed algorithm is based on the combination of Least Square Support Vector Machine and Group Method of Data Handling. The performance of the proposed algorithm was compared with those two methods, using different strategies (Direct, Recursive and DirRec). The results showed proposed algorithm with DirRec strategy has a significant improvement over those two methods and traditional ANN. The effects of the aerosol data, water vapor data and ozone content data on the output of the clear-sky models were studied for estimation of clear-sky solar irradiance [14]. In this study, the performance of three clear-sky solar irradiance models, namely European Solar Radiation Atlas clear sky model, simplified SOLar Irradiance Scheme clear sky model, and Reference Evaluation on Solar Transmittance 2 clear sky model, was evaluated and compared. In addition, they studied the performance of those models using the same atmospheric input data but at different elevation. In [15] the effect of uncertainty in temperature is considered in load forecasting; however, this work was only on energy demand and the only weather variable in this study was temperature. Chen et al. [16] proposed an artificial intelligent based technique for forecasting solar power which required the past power measurement, solar irradiance forecast, humidity and temperature as inputs. These studies have laid a solid ground for forecasting the generations of renewable DERs and at the same time revealed the needs of considering the impact of weather forecasting errors on the forecasted power outputs.

Fig. 1 shows a typical process for the solar PV generation forecasting using weather data and historical generation data.

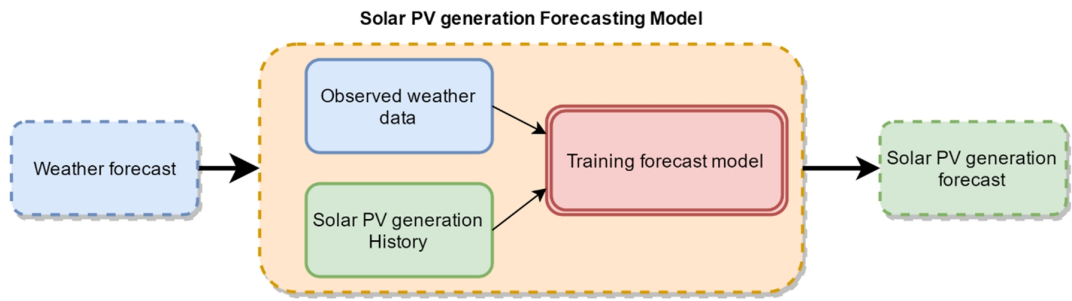

*Fig. 1. Typical solar PV generation forecasting process*

As shown, the forecasting model is trained using observed weather data and historical data of solar PV generation. After the model is well trained using artificial neural network (ANN), the forecasted weather data will be used to have solar PV generation forecast. The major contribution of this study is to analysis the uncertainty of weather forecast and its effect on the solar PV generation forecast. In this study, the impact of weather forecast errors for several weather variables including sky cover, dew point, relative humidity, temperature, and wind on the performance of solar PV generation forecasting is assessed.

As a case study, the solar energy generated by the solar PV panels on the rooftop of Engineering and Science buildings at Binghamton University is considered. Accordingly, the most commonly used weather variables in solar forecasting such a sky cover, dew point, relative humidity, temperature, and wind are considered for assessment. The aforementioned variables for both observed and forecasted values of 6 days ahead are collected. The errors in the forecasted weather variables of each day are evaluated by statistical metrics. Using the bootstrapping method, the uncertainty of the forecasting errors is quantified. Then, by applying correlation analysis, the influential variables are identified and selected for forecasting. Finally, by applying the forecasted weather variables along with the observed data, the performance of forecasting models in dealing with inputs errors is studied.

The rest of the paper is organized as follows. In Section 2, observed and forecasted weather data and their classification are discussed. Section 3 elaborates the weather data error analysis and quantifies errors in forecasted variables. Simulation results are presented in section 4 where the influential weather variables are identified using a correlation method and the performance of the energy forecasting method with observed and forecasted weather data is evaluated. At the end, the conclusions are drawn in Section 5.

## 2. Observed and Forecasted Weather Data Acquisition

To assess the effects of weather variables in forecasting model, weather data are extracted from National Oceanic and Atmospheric Administration (NOAA), which provides data in public domain. Both forecasted and observed data of weather variables are available for most of local areas in US with hourly resolution. The observed weather variables are available online for four days ahead and most of them are update hourly while the forecasted weather variables are updated hourly for six days ahead.

The most influential driving variables in solar PV generation forecasting model are time, date, and weather variables [10] [17]. For selecting predictors in solar energy forecasting model, sky cover, relative humidity, dew point, temperature, wind speed, pressure, and precipitation are usually considered as weather variables. The observed weather variables provided by the NOAA include weather condition, sky cover, dew point, relative humidity, visibility, pressure, temperature, precipitation, and wind speed. However, the forecasted weather variables provided by the NOAA are sky cover, dew point, relative humidity, participation potential, relative humidity, temperature, and wind. Among observed and forecasted categories, sky cover, dew point, relative humidity, temperature, and wind speed are common in both categories and also represented in hourly interval. Accordingly, the aforementioned variables are considered for weather data analysis and selecting predictors in the following sections.

However, the sky cover, which is one of the most important driving inputs in solar PV generation forecasting, is represented differently for the observed and forecasted data. In the observed weather data, the sky cover is represented by categories as depicted in Table 1 while in the forecasted data, it is represented by numerical percentage. To compare the observed and forecasted data of this weather variable, both data types are categorized in a common category. The fourth column of Table 1 gives a numerical percentage category for the observed data. Similarly, the forecasted sky cover data is also classified within five groups shown in the fourth column of Table 1.

**Table 1** Numerical classification for observed and forecasted sky cover

| Sky Condition | Opaque Cloud Coverage | Opaque Cloud Coverage (%) | Percentage Category (%) |
|---|---|---|---|
| Clear | 1/8 and less | Sky Cover < 12.5 | 0 |
| Mostly Clear | 1/8 to 3/8 | $12.5 \leq$ Sky Cover $< 37.5$ | 25 |
| Partly Cloudy | 3/8 to 5/8 | $37.5 \leq$ Sky Cover $< 62.5$ | 50 |
| Mostly Cloudy | 5/8 to 7/8 | $62.5 \leq$ Sky Cover $< 87.5$ | 75 |
| Cloudy | 7/8 to 8/8 | $87.5 \leq$ Sky Cover | 100 |

## 3. Weather Data Analysis

As mentioned, the data provided by the NOAA for the historical forecasted data spans for 6 days ahead. To assess the error corresponding to weather forecasting, the observed weather data are compared with historical forecasted weather data for a complete year during May 20th, 2016 to the end of the day on May 19th, 2017 with error metrics.

To quantify errors, there are different commonly used metrics such as the mean absolute percentage error (MAPE), MAE, mean squared error (MSE), and root-mean-square error (RMSE) [18]. In this study, the MAPE defined by (1) is used to evaluate the error in solar PV generation forecasting results. However, since some weather indicators are zero, the MAE defined by (2) is used to represent the error in weather forecasting.

$$\text{MAPE} = \frac{1}{N} \sum_{i=0}^{N} \left| \frac{y_i - f(x_i)}{y_i} \right| \times 100 \quad (1)$$

$$\text{MAE} = \frac{1}{N} \sum_{i=0}^{N} |y_i - f(x_i)| \quad (2)$$

Where, $N$ is the number of observations, $y_i$ is the actual target value at time instant $i$, the symbol $x_i$ is the input vector, and $f$ is the forecasting model.

In addition, to estimate the error statistics, the bootstrapping method is applied [19]. The bootstrapping is an efficient numerical approach for estimating some statistical parameters like mean and standard deviation of population from a sample. Bootstrapping, which is based on resampling and replacement of a sample, does not make any assumptions about the distribution of the sample data. However, it requires that the sample data and its size should be sufficient to well represent the population distribution.

In addition, bootstrapping can be used to derive the uncertainty of the estimated statistical parameters. Such uncertainty represented by confidence intervals (CIs) claims to cover the true statistics of population within the intervals with a specified probability. For example, in [20], such a specified probability is 95% which means that the true value of population is located in CIs with the probability of 0.95. For this case, the number of bootstrap resampling cycles and the probability of the CIs are 2500 and 95%, respectively [21].

In this study, the forecast errors, which are the difference between observed and forecasted weather data, are calculated using MAEs in (2) for each hour of 6 successive days. In addition, to consider the likely direction of forecasting error, bias in error defined by (3) is also considered in the analysis. The sign of the bias represents the direction of the errors, where positive bias indicates the observed data is more than the forecasted value and vice versa.

$$\text{Bias} = \frac{1}{N} \sum_{i=0}^{N} (y_i - \hat{y}_i) \quad (3)$$

Where $\hat{y}_i$ is the forecasted variable which in this case is the historical forecasted weather variables and $y_i$ is the observed weather variables. In addition, in the calculation of MAE in (2), $\hat{y}_i$ is also used for $f(x_i)$ and observed weather variables are used for $y_i$. To imply the population statistics of the aforementioned error metrics, bootstrap method is applied. Table 2 depicts the results of calculation for each weather variable in each day of forecasting. On the first row, each day is shown as D #no.

**Table 2.** Statistics of error in weather forecasting for 6 days Ahead

| Type | Statistics | D #1 | D #2 | D #3 | D #4 | D #5 | D #6 |
|------|-----------|------|------|------|------|------|------|
| SC   | Bias      | -6.46 | -5.49 | -.4.49 | -.221 | -2.2 | -2.1 |
|      | MAE       | 24.27 | 25.74 | 28.44 | 31.32 | 33.8 | 35.57 |
| DP   | Bias      | -2.22 | -2.56 | -2.8 | -2.7 | -2.73 | -2.68 |
|      | MAE       | 2.93 | 3.44 | 3.8 | 4.13 | 4.7 | 5.32 |
| RH   | Bias      | 0.5 | -0.34 | -0.9 | -0.82 | -0.92 | -0.9 |
|      | MAE       | 8.88 | 9.88 | 10.64 | 11.11 | 11.83 | 12.27 |
| T    | Bias      | -2.1 | -2.06 | -2.07 | -2 | -1.94 | -1.88 |
|      | MAE       | 3.08 | 3.26 | 3.63 | 3.95 | 4.46 | 4.97 |
| W    | Bias      | 1.74 | 2.07 | 2.67 | 3.35 | 3.4 | 3.25 |
|      | MAE       | 3.09 | 3.44 | 3.78 | 4.22 | 4.36 | 4.37 |

SC: sky cover; DP: dew point; RH: relative humidity; T: temperature, W: wind.

Table 2 shows that the bias takes negative values for almost all of the forecasted days except for wind which is totally vice versa. This result indicates that the NOAA generally forecasts a value more than real observed variables for sky cover, dew point, relative humidity, and temperature whereas for the wind, the NOAA provides underestimated values. Such biases in weather forecasting may result in overestimation and underestimation in energy forecasting. In addition, the results of bias indicate that the forecasting residuals in NOAA forecasting model do not have a zero mean. When the residuals of a forecasting model have a mean other than zero, the forecasting model is biased and it can be improved to have better results [22].

The results of autocorrelation function (ACF) also shows inefficacy of the weather forecast model. As an example, Fig. 2 shows the ACF of the sky cover forecast error for one day ahead with the lag length of 100. As it is illustrated, there are more than 5% of the spikes out of the bounds. The results of ACFs for other weather variables also show similar results. Therefore, the weather forecasting model in NOAA violates that assumption of no autocorrelation in the residuals which means there is more information left over which can be implemented to improve the results of forecasting.

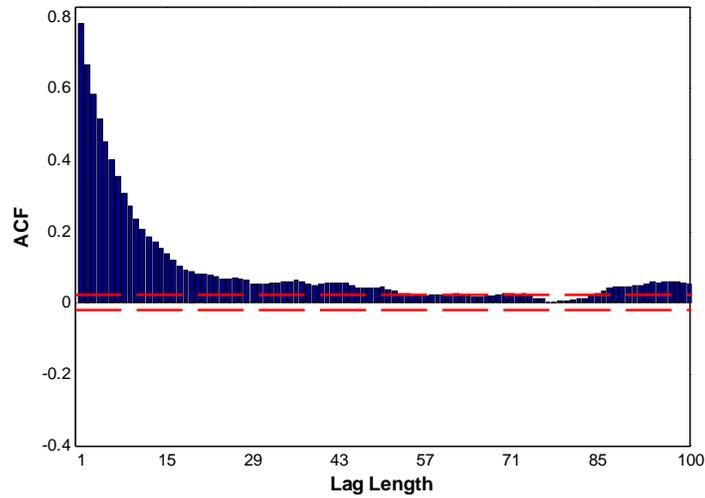

*Fig. 2. ACF of the forecasted sky cover for one day ahead*

In Table 3 the results of 95% CIs for the MAE of the errors is depicted. The two numbers in the table cells are the lower and upper bounds, respectively.

**Table 3.** 95% CIs of MAE in forecasting of weather variables (lower bound, upper bound)

| Type | D #1 | D #2 | D #3 | D #4 | D #5 | D #6 |
|---|---|---|---|---|---|---|
| SC | 22.6, 25.9 | 24.1, 27.37 | 26.9, 29.95 | 29.87, 32.77 | 32.36, 35.22 | 34, 37 |
| DP | 2.7, 3.1 | 3.23, 3.64 | 3.57, 4.03 | 3.86, 4.38 | 4.4, 5 | 5, 5.64 |
| RH | 8.4, 9.4 | 9.3, 10.4 | 10, 11.2 | 10.45, 11.8 | 11.15, 12.52 | 11.6, 12.95 |
| T | 2.9, 3.3 | 3.1, 3.4 | 3.4, 3.85 | 3.7, 4.1 | 4.2, 4.7 | 4.6, 5.3 |
| W | 2.93, 3.26 | 3.26, 3.62 | 3.59, 3.98 | 4.02, 4.43 | 4.14, 4.59 | 4.14, 4.6 |

Another representation of the MAE in Table 2 and CIs in Table 3 is illustrated by Figs. 3 and 4 where Fig. 3 represents MAE of sky cover and similarly Fig. 4 illustrated MAEs in other weather variables.

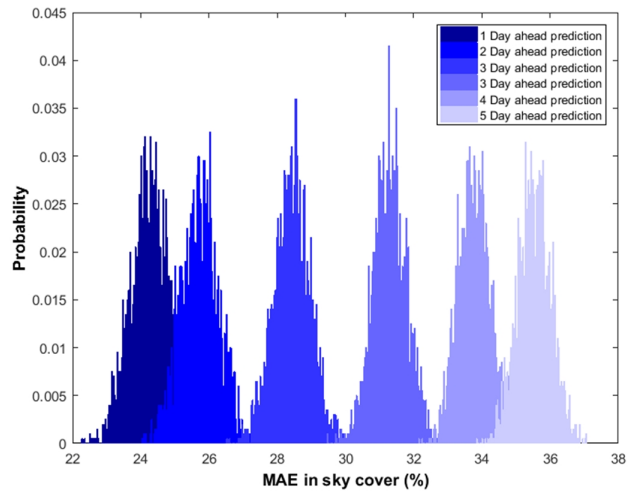

*Fig. 3. MAE of forecasted sky cover for six days ahead*

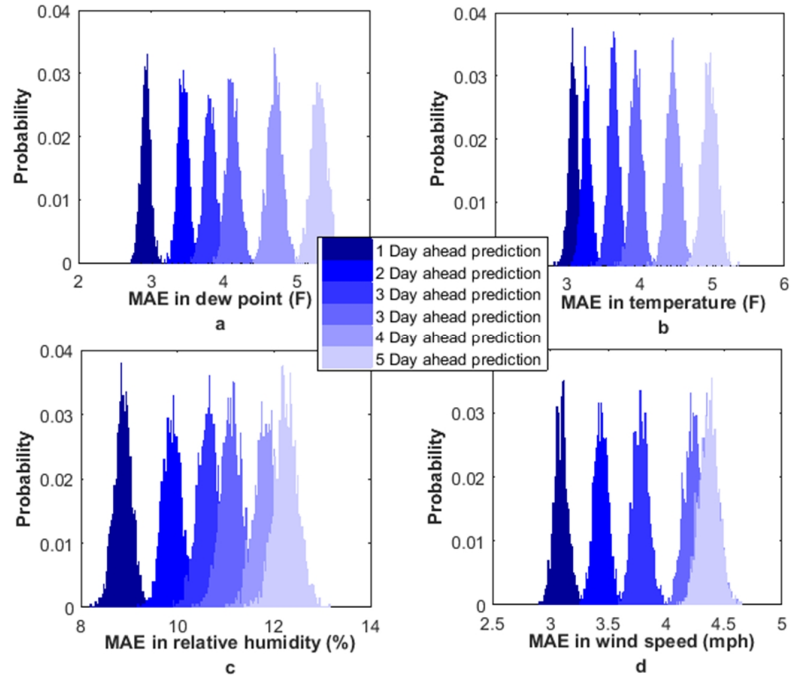

*Fig. 4.* *MAEs of forecasted variables six days ahead*
a MAE of forecasted dew point for six days ahead
b MAE of forecasted temperature for six days ahead
c MAE of forecasted relative humidity for six days ahead
d MAE of forecasted wind speed for six days ahead

## 4. Simulation Results

To study the effect of observed and historical weather data on solar PV energy generation, the solar PV generation by the solar panels installed on the rooftop of the Engineering and Science building at the State University of New York at Binghamton is considered as case study. For this purpose, the observed weather data and corresponding 6-day ahead forecasted weather data for the likely influential weather variables on energy forecasting and at the same location of solar panels are considered during one study year from May 20$^{th}$, 2016 to end of the day on May 19$^{th}$, 2017. In addition, the solar energy generated by the panels are acquired by hourly resolution during the studying time. Fig. 5 illustrates the daily solar energy for the case study during studying year. The solar panels of the case study are capable of providing maximum nominal energy of 120 kWh.

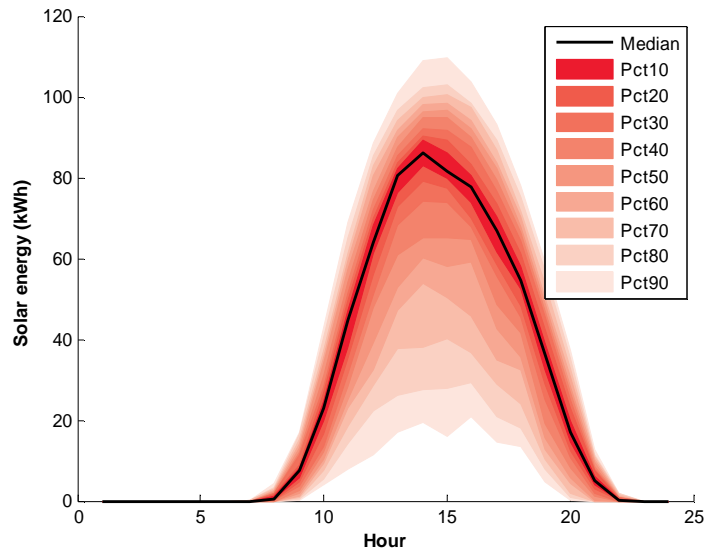

*Fig. 5. 24-hour profile of solar PV generation for the case study*

For the purpose of forecasting analysis, the peak energy is selected and both observed and forecasted weather variables corresponding to the peak time of solar generation are extracted from weather data sets. Thus, from this point on, solar energy refers to the daily peak solar energy generation and weather data refers to the corresponding data at daily peak time of solar energy generation.

In most cases, weather variables along with other indicators like time and date are used as predictor variables in solar energy forecasting models. As mentioned in second section, the available variables in both observed and forecasted weather data provided by the NOAA are sky cover, dew point, relative humidity, temperature, and wind speed. However, the existence unrelated variables as predictors in training model may lead to inaccurate results. In addition, the correlation between predictors may lead to misleading results. Thus, a predictors selection analysis is conducted before applying the candidate predictors in forecasting model. Note that the weather data applied in the training process of energy forecasting are observed data and here in this case also observed weather variables are applied in predictors analysis.

Table. 4 shows the correlation analysis between solar energy and five weather variables. As shown in this table, the correlation coefficients between energy and five weather variables indicate that the dew point and wind speed have very low correlations with energy. On the other hand, relative humidity has the highest correlation with energy with the value of - 0.61. In addition, dew point and temperature are highly correlated (r= 0.9). Fig. 6 illustrates the scatter plots of these four variable of energy-dew point, energy-wind, energy-relative humidity, and dew point-temperature. As shown in Fig. 6. a and Fig. 6. b, the scatter plots of energy with dew point and wind have sporadic patterns while as illustrated in Fig. 6. c, energy and relative humidity are highly correlated. In addition, the scatter plot of dew point and temperature, shown in Fig. 6. d, indicates high correlation between these two predictors. Thus, considering the results of correlation analysis, dew point and wind are excluded from the predictors set and three predictors of sky cover, relative humidity, and wind are selected for training the solar PV energy forecasting. Note that the authors also considered all states of possible combination of predictors (31 states) in training and validation of forecasting model using artificial neural network (ANN) and the results also confirmed the optimal selection of predictors as sky cover, relative humidity, and temperature.

**Table 4.** Results of correlation between energy and weather variables

|  | **Energy** | **Sky cover** | **Dew point** | **Relative humidity** | **Temperature** | **Wind** |
|---|---|---|---|---|---|---|
| **Energy** | 1 | -0.42 | 0.18 | -0.61 | 0.44 | -0.09 |
| **Sky cover** | -0.42 | 1 | 0.14 | 0.35 | -0.09 | 0.14 |
| **Dew point** | 0.18 | 0.14 | 1 | 0.28 | 0.9 | -0.17 |
| **Relative humidity** | -0.61 | 0.35 | 0.28 | 1 | -0.15 | -0.02 |
| **Temperature** | 0.44 | -0.09 | 0.9 | -0.15 | 1 | 0.17 |
| **Wind** | -0.09 | 0.14 | -0.17 | -0.02 | 0.17 | 1 |

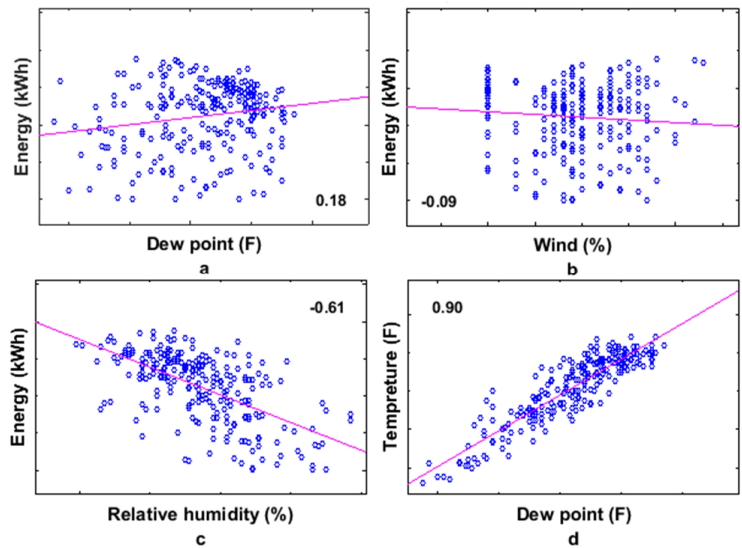

*Fig. 6. Correlation between solar energy and weather variables*
a correlation between energy and dew point
b correlation between energy and wind
c correlation between energy and relative humidity
d correlation between temperature and dew point

As mentioned before, there are many methods and models for training and modeling for solar energy forecasting. However, since the purpose of this study is more about the analysis of weather variables in solar PV energy forecasting and the effect of their uncertainties, only one of the forecasting methods is chosen for the simulation. Among lot of the forecasting methods, ANN method, is one of the most commonly used and efficient methods and it is applied for this case [23]. The ANN method is like a black-box model, which provides an efficient way to model a complex nonlinear system. To model a system using an ANN model, there is no need to figure out the closed-form equations of the system or to know the complex relationship between input and output variables [24]. The neural network used in this study is a feed forward supervised learning model with one hidden layer. The number of hidden neurons in the hidden layer is three [25]. In addition, the Levenberg-Marquardt algorithm is used for model training.

The solar PV generation forecasting model sets the daily peak value of solar energy as the dependent variable and three selected weather indicators of sky cover, relative humidity, temperature at the corresponding peak time as independent variables. First, the model is trained with solar energy and observed weather data. Then, the historical forecasted weather data for 6 days ahead are applied to the trained model. Accordingly, there are 6 forecasted solar PV energy sets corresponding to the 6 days of forecasted weather data as inputs. The forecasted energy for 6 days ahead are compared with the real generated solar energy and corresponding errors for each day are represented by MAPE and MAE. Fig. 7 depicts the results of errors for the observed and historical forecasted weather data. The blue bars show the errors corresponding to the MAPEs of forecasted energy using historical forecasted data and the red bar is corresponding to the MAPE of forecasted energy using observed weather data. Based on the increasing error trend, a huge proportion of error belongs to the observed data. However, in the forecasting with historical forecasted data (which carry significant input errors) the error is not significant in compared with the observed data (which is considered as actual weather data with likely small error). Table 5 also depicts the result of MAE for the forecast errors using observed and historical forecasting data.

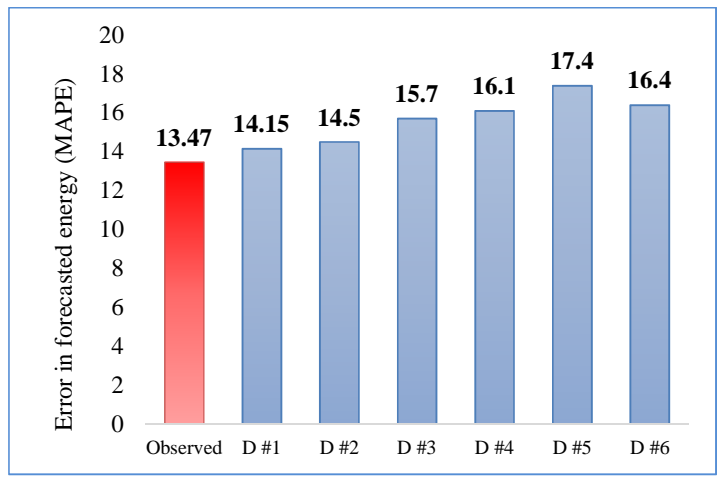

*Fig. 7. MAPEs results using observed and historical forecasted data*

**Table 5.** Results of applying forecasted weather variables in solar energy model

| Statistics | D #1 | D #2 | D #3 | D #4 | D #5 | D #6 |
|---|---|---|---|---|---|---|
| *MAE* | 10.29 | 10.74 | 11.71 | 11.98 | 13 | 12.1 |

According to the Fig. 7 and Table 5, the ANN model is robust enough to handle the input errors. Thus, the forecast weather variables can be handled with a well-designed forecasting model although they may carry considerable errors. This result can help a network manager to efficiently estimate DERs hosting capacity considering by defining an acceptable errors dictated by forecasted weather data.

Although energy forecasting model can handle the error in forecasting weather variables, such an error can be decreased if the influential weather variables are identified and weather forecaster improves the accuracy of those influential variables, specifically. To achieve this purpose, a sensitivity test is conducted in which three scenarios for the three weather variables inputs (sky cover, relative humidity, and temperature) of energy forecasting model are implemented. In each scenario, it is supposed that for one of the weather variables, the actual weather data is available for the 6 days ahead instead of forecasted data while for other weather variables, historical forecasted data are applied in energy forecasting model. Accordingly, in each scenario, the observed data for one of weather variables along with historical data for other weather variables are applied as inputs of the energy forecasting model and the MAPEs of all scenarios are compared with the MAPE of forecasting using only forecasted weather variables. The result of sensitivity test for all three scenarios along with forecasting using only forecasted weather variables is shown in Fig. 8. In this figure, the MAPE illustrated by continuous black line is the error in energy forecasting model using only historical forecasted weather variables.

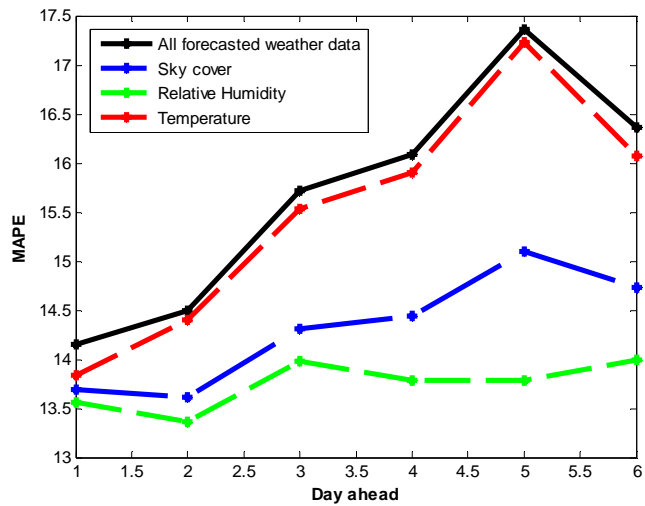

***Fig. 8.*** *Sensitivity analysis of weather variables*

As seen in Fig. 8, the green line which belongs to the scenario test of relative humidity has the least MAPE for the energy forecasting. In this scenario, only for the relative humidity, the observed data of 6 days ahead are applied in energy forecasting model while other weather variables are retained as forecasted values. Accordingly, if the forecaster can improve the accuracy of this weather variable by using accurate measurement and/or weather forecasting model, the energy forecaster will be able to provide better forecasting results. In addition, the blue line, which represents sky cover variable, also can improve the results of energy forecasting if the forecasted value of this variable is provided accurately.

## 5. Conclusion

In this paper, the uncertainty in weather forecasting was studied for solar PV generation forecasting. For this purpose, the data for both observed and 6-day forecasted values of weather variables were extracted from NOAA for a complete year of studying time. The common variables between both observed and forecasted variables were selected as potential influential predictors for solar energy forecasting model. Then the errors in weather forecasting were derived by comparing the observed and 6-day forecasted values. Using the bootstrapping method, the errors corresponding to each 6 day of forecasting were presented by statistical metrics. The results of error analysis indicate bias and overestimating in weather forecasting for all weather variables except wind speed which is underestimated for all 6 days of weather forecasting. Using correlation analysis, the most influential variables i.e. sky cover, relative humidity, and temperature were selected for energy forecasting model training. The impact of forecasted weather data errors on the forecasting model was assessed using the ANN model. The MAPEs results show that although there are significant errors in historical forecasted data, the forecasting model can handle the errors, decently. Finally, a sensitivity test on weather variable was performed to identify weather variables whose accurate values can significantly improve the energy forecasting. The results show that relative humidity plays the most influential role in energy forecasting and an energy forecaster can significantly increase the accuracy of their results during 6 days ahead by having accurate forecast of relative humidity.